\documentclass{article}

\usepackage{microtype}
\usepackage{graphicx}
\usepackage{subfigure}
\usepackage{booktabs} 

\usepackage[colorlinks=true, allcolors=blue]{hyperref}

\usepackage[accepted]{icml2025}

\usepackage{amsmath}
\usepackage{amssymb}
\usepackage{mathtools}
\usepackage{amsthm}
\usepackage{rotating}
\usepackage{multirow}
\usepackage{makecell}
\usepackage[utf8]{inputenc} 
\usepackage[T1]{fontenc}    
\usepackage{xcolor,colortbl}
\usepackage[colorlinks=true, allcolors=blue]{hyperref}      
\usepackage{booktabs}       
\usepackage{amsfonts}       
\usepackage{nicefrac}       
\usepackage{microtype}      
\usepackage{natbib}
\usepackage{microtype}
\usepackage{graphicx}
\usepackage{subfigure}
\usepackage{booktabs} 
\usepackage{epigraph}
\usepackage{pifont}
\usepackage{amsmath}
\usepackage{amssymb}
\usepackage{mathtools}
\usepackage{amsthm}
\usepackage{xurl}
\usepackage{contour}
\usepackage{mathtools} 
\usepackage{svg}
\usepackage[commandnameprefix=ifneeded]{changes}
\usepackage{float}
\usepackage{bbm}

\newcommand{\stkout}[1]{\ifmmode\text{\sout{\ensuremath{#1}}}\else\sout{#1}\fi}

\setdeletedmarkup{\stkout{#1}}
\usepackage[capitalize,noabbrev]{cleveref}
\usepackage{enumitem}
\usepackage[capitalize,noabbrev]{cleveref}

\theoremstyle{plain}

\theoremstyle{definition}

\theoremstyle{remark}

\definecolor{teal1}{HTML}{b5d1ae}
\definecolor{teal2}{HTML}{80ae9a}
\definecolor{teal3}{HTML}{568b87}
\definecolor{teal4}{HTML}{326b77}
\definecolor{teal5}{HTML}{1b485e}
\definecolor{teal6}{HTML}{122740}
\definecolor{darkred}{HTML}{ae282c}
\definecolor{red}{HTML}{e02b35}
\definecolor{medred}{HTML}{d47264}
\definecolor{darkblue}{HTML}{2066a8}
\definecolor{blue}{HTML}{1a80bb}
\definecolor{medblue}{HTML}{3594cc}
\definecolor{lightblue}{HTML}{8cc5e3}

\newcommand{\myrotcell}[1]{\rotcell{\makebox[0pt][l]{#1}}}

\def\iid.{\mbox{i.}\mbox{i.}\mbox{d.}}
\def\ood.{\mbox{o.}\mbox{o.}\mbox{d.}}
\def\eg.{\mbox{e.}\mbox{g.}}
\def\ie.{\mbox{i.}\mbox{e.}}

\long\def\LB#1{\textcolor{black}{#1}}

\newenvironment{itemize*}%
  {\begin{itemize}%
    \vspace{-\topsep}%
    \setlength{\itemsep}{0pt}}
  { \vspace{-\topsep}%
   \end{itemize}}

\icmltitlerunning{Finetuning with Very-large Dropout}

\begin{document}

\twocolumn[
\icmltitle{Finetuning with Very-large Dropout}



\icmlsetsymbol{equal}{*}

\begin{icmlauthorlist}
\icmlauthor{Jianyu Zhang}{nyu,fair}
\icmlauthor{L\'eon Bottou}{fair,nyu}

\end{icmlauthorlist}

\icmlaffiliation{nyu}{New York University, New York, NY, USA.}
\icmlaffiliation{fair}{FAIR, Meta, New York, NY, USA}

\icmlcorrespondingauthor{Jianyu Zhang}{jianyu@nyu.edu}

\icmlkeywords{Machine Learning, ICML}

\vskip 0.3in
]



\printAffiliationsAndNotice{}  

\begin{abstract}

It is impossible today to pretend that the practice of machine learning is always compatible with the idea that training and testing data follow the same distribution. Several authors have recently used ensemble techniques to show how scenarios involving multiple data distributions are best served by representations that are both richer than those obtained by regularizing for the best in-distribution performance, and richer than those obtained under the influence of the implicit sparsity bias of common stochastic gradient procedures.

This contribution investigates the use of very high dropout rates instead of ensembles to obtain such rich representations. Although training a deep network from scratch using such dropout rates is virtually impossible, fine-tuning a large pre-trained model under such conditions is not only possible but also achieves out-of-distribution performances that exceed those of both ensembles and weight averaging methods such as model soups. 

This result has practical significance because the importance of the fine-tuning scenario has considerably grown in recent years. This result also provides interesting insights on the nature of rich representations and on the intrinsically linear nature of fine-tuning a large network using a comparatively small dataset.

\end{abstract}

\section{Introduction}
\label{sec:introduction}

The practice of machine learning has been shaped by the assumption that training and testing examples are independently drawn from the same unknown probability distribution. \textit{This is seldom the case in modern settings, not only because this \iid. assumption breaks down for the problems of interest, but also because it is often convenient to use multiple datasets that are known to follow different distributions.} For instance, we may pre-train a deep network on a large dataset, fine-tune it on a smaller dataset specific to the task of interest, and test on a collection of tasks designed to benchmark various aspects of the system.

Many of the tenets of machine learning should therefore be regarded with healthy suspicion. For instance, under the \iid. assumption, favoring solutions with sparse representations has well-known benefits on the generalization performance. Yet, several authors \citep{zhang2022rich,zhang2023learning,chen2023towards} make the point that scenarios involving multiple distributions are best served by ``\emph{richer representations}'' that contain redundant features, that is, features that do not improve the model performance on the training distribution, but could prove helpful when the distribution changes.

It would be nice to construct such rich representations by merely optimizing the expectation of a suitable loss function for a single training distribution, for instance using stochastic gradient techniques.
Alas, this hope is contradicted by the implicit sparsity bias of stochastic gradient algorithms \citep{andriushchenko2023sgd,blanc2020implicit}. In a nutshell, a feature only survives when it brings an incremental training error advantage relative to what can be achieved using all the other features already present in the network. \LB{We slightly abuse the terminology and call them ``strongly relevant''. However, features that are not strongly relevant might nevertheless 
\begin{itemize*}
    \item[\textbf{(a)}] be incrementally useful when the data follows a different distributions of interest, or
    \item[\textbf{(b)}] be useful under the training distribution when added to certain subsets of the other existing features instead of all of them (``weakly relevant'').
\end{itemize*}}
It is therefore tempting to ``enrich" the representation with features of type (b), which can be found using the training data, and hope that some of these will turn out to also be features of type (a) whose inclusion helps when the data distribution changes.

The dropout technique \citep{srivastava2014dropout} seems well suited to find \LB{weakly relevant features} because randomly masking units of a representation layer during training amounts to forming random subsets of all other available features. However, in order to form small subsets, one would have to use very high levels of dropout. Unfortunately, training a sizable deep network from scratch with such a large dropout is practically impossible. \LB{Instead, computationally demanding methods, such as adversarial sampling \citep{zhang2022rich,chen2023towards} and representation ensembles \citep{zhang2023learning}, have been proposed to find \LB{weakly relevant features} while training a network from scratch.}

There is however a practically meaningful scenario \LB{in which we can use an extremely aggressive dropout: fine-tuning a pre-trained network using a comparatively small dataset.  This is possible because such a fine-tuning operation makes only modest changes to the network weights.} For example, several authors \citep{rame2022diverse, wortsman2022model} argue that fine-tuned networks remain ``\textit{linearly connected}'', that is averaging the parameters of multiple fine-tuned networks approximate the ensemble of these networks.  \citet{evci2022head2toe} even show that a linear classifier on top of the union of internal-layer features of pre-trained residual networks can match or exceed the performance of fine-tuning.

In the present work, we adopt the out-of-distribution fine-tuning setup (\textit{three-distributions}) of \citet{rame2022diverse}. In this framework, we have access to a model pre-trained using a large dataset for a task weakly related to the task of interest. This pre-trained model is then fine-tuned on datasets that illustrate the task of interest, and then tested on a dataset for the same task but with a different distribution.
However, instead of enriching the representations by constructing ensembles \citep{zhang2023learning} or averaging weights \citep{rame2022diverse,rame2022recycling,wortsman2022robust}, we simply \emph{fine-tune using very large dropout levels}, randomly masking \LB{above} 90\% of the units in the representation layer. We find that this simple approach \emph{exceeds the performance of both ensemble and weight-averaging methods}. This result is not only \emph{practically meaningful}, but also clarifies the idea of \emph{richer representation}.

\section{Related Work}
\label{sec:relate_work}

\paragraph{Constructing versatile representations}

Reusing or transferring features across related tasks has been commonplace for more than one decade \citep{collobert-2011, tr-bottou-2011, sharif2014cnn} and plays a fundamental role in the appeal of foundational models \citep{bommasani-2021}. However, once the optimization process has identified a set of features that is sufficient to achieve near-optimal performance on the training set, additional features are often discarded because they do not bring an incremental benefit to the training error, despite the fact that they may independently carry useful information \citep{zhang2023learning}. 

Researchers have devised ways to obtain more versatile representations by engineering a diversity of datasets, architectures, and even hyper-parameters \citep{chen2020simple, wang2022cross, dvornik2020selecting, bilen2017universal, gontijo2021no, li2021universal, li2022universal, chowdhury2021few}, as an alternative to the most popular approach which consists of simply using ever larger datasets \citep{bommasani2021opportunities}. 

Interesting results have also been obtained without engineering diversity and without increasing the dataset sizes. \citet{zhang2022rich} and \citet{chen2023towards} propose to discover rich representation through multiple training episodes that adversarially reweigh the training dataset to impede the use of previously learned features. \citet{zhang2023learning} show that surprisingly good results can be obtained by concatenating the representations of multiple networks that are trained in exactly the same way, save for the random seed used in the stochastic gradient process.

\paragraph{Fine-tuning as a near-linear process}

Although modern deep residual networks feature highly complex nonconvex cost functions, several authors have shown that their final training phase remains mostly confined to a nearly-convex attraction basin \citep{izmailov2018averaging, li2018visualizing, frankle2020linear}. The same observation holds when fine-tuning a large pre-trained network using a dataset whose size is considerably smaller than the dataset size one would need to train such a large network from scratch.  As long as one starts from the same pre-trained model, \citet{wortsman2022model} and \citet{rame2022diverse, rame2022recycling} observe that averaging the weights of diverse fine-tuned models can reproduce the \iid. and \ood. performances of the ensemble of these models, implying that fine-tuning is a near-linear process.

\citet{maddox2021fast} and \citet{mu2019gradients} propose instead to approximate the fine-tuning process with a first-order Taylor expansion, obtaining a linear system operating on top of the \textsc{ntk} features.  \citet{evci2022head2toe} match the performance of fine-tuning by merely learning a strongly regularized linear model that takes all internal layer states as inputs. Meanwhile, \cite{yu2023low} efficiently fine-tune large foundational language models by essentially restricting the weight updates to low dimensional manifolds.

\paragraph{Fine-tuning with very large dropout}

Our contribution \LB{advocates using very large dropout in the fine-tuning scenario in order to force the learning algorithm to create a redundant representation without specifically engineering diversity. We do not seek to propose new dropout variations \citep{dna}, understand dropout from either an overfitting/underfitting perspective \citep{liu2023dropout} or from a Bayesian perspective \citep{gal2016dropout}.}

\section{Fine-tuning and dropout}
\label{sec:method}


\subsection{The three-distributions setup}

The \textbf{\textit{two-distributions}} setup is commonly used for transfer learning. In this setup, features $\Psi$ are obtained by pre-training a network on a large training set associated with a first distribution $\mathcal{T}_{\text{p}}$. These features are then used to construct or initialize a new model $\omega_{\text{d}} \circ \Psi$, which is then trained using a smaller training set associated with a second distribution $\mathcal{T}_{\text{d}}$. The question is to determine which pre-training approach is most likely to make the features $\Psi$ useful for the transfer task $\mathcal{T}_{\text{d}}$.

The \textbf{\textit{three-distributions}} setup \citep{rame2022diverse} views the pre-trained model as a base model that is assumed very rich but whose training process is beyond our control (e.g., a fundational model). The features $\Psi$ of the pre-trained model are then incorporated into a new model $\omega_{\text{d}} \circ \Psi$ that is fine-tuned using a second distribution $\mathcal{T}_{\text{d}}$ and eventually tested on a third distribution $\Tilde{\mathcal{T}}_{\text{d}}$ illustrating the same general task as the second distribution (e.g., using the same classification labels.) The question is then to determine which fine-tuning approach is most likely to produce a model that will perform robustly under the eventual testing distribution $\Tilde{\mathcal{T}}_{\text{d}}$.

\subsection{Examples}

Considering a logistic regression with parameter $\omega\in\mathbb{R}^n$ operating on a vector $\Psi$ of $n$ features and predicting a binary target $Y$ representing our ${\mathcal{T}}_{\text{d}}$ distribution. Assume further that each individual feature $\Psi_i, i \in [1,\dots, n]$ perfectly predicts $Y$, that is, zero classification error can be achieved with a regression $\omega$ whose only nonzero parameter marks the $i$-th feature. During gradient-based optimization, achieving zero loss by using only one feature prevents the system from using the other features, because of the ``gradient starving'' phenomenon \citep{pezeshki2021gradient}.   
We now evaluate this trained system on a target distribution $\Tilde{\mathcal{T}}_{\text{d}}$ that only differs from ${\mathcal{T}}_{\text{d}}$ because some features were missing and have been replaced by zeroes. If our trained system (on ${\mathcal{T}}_{\text{d}}$) depends only on one feature, we better hope that this is not one of the missing ones in target distribution $\Tilde{\mathcal{T}}_{\text{d}}$.

\textbf{In this linear case}, the following three strategies are equivalent in terms of encouraging the optimization process to learn more features: 1) feature-bagging (ensemble) \cite{bryll2003attribute}; 2) Dropout; 3) $L_2$ regularization on $\omega$ (Check \citet{srivastava2014dropout} for the proof). We know that the feature-bagging approach solves the problem above by construction. Thus, in the linear case, all three strategies solve the above problem.

\textbf{In the case of a multilayer network}, however, this equivalence is broken. In particular, $L_2$ regularization on the inner layer parameters plays the different role of encouraging sparse representations \cite{blanc2020implicit, andriushchenko2023sgd}. Dropout and deep ensembles may achieve comparable error rates in distribution but differ sharply when it comes to estimating prediction uncertainty \citep{ashukha2020pitfalls}. These differences become very important when one fine-tunes the model using out-of-distribution data, making deep ensembles and weight averaging ensembles more attractive than dropout for \ood. generalization \cite{rame2022diverse, rame2022recycling, wortsman2022model, swad, wa}. 

Our contribution shows that using a very large dropout rate during fine-tuning (rather than during initial training) substantially improves on the \ood. performance of both ensemble and weight-averaging. This simple approach was not considered before, possibly because such large dropout rates are not usable during pre-training, resulting in poor performance overall.

\subsection{Method}

The key results described later in this paper have been obtained with a very simple method.  The base model is a deep learning network with residual connections trained on data $\mathcal{T}_{\text{p}}$ that is related to but substantially larger than the datasets illustrating the task of interest. Some of these datasets ($\mathcal{T}_{d}$) are used to fine-tune the base model. Performance is reported on both held-out data from the fine-tuning datasets (\iid. performance on $\mathcal{T}_{d}$) and data from the remaining datasets (\ood. performance on $\Tilde{\mathcal{T}}_{d}$). 

\LB{We focus on residual networks because fine-tuning has been found to hardly change the inner layers of non-residual networks (\citealp{raghu2019rapid}, fig 2). 
In contrast, skip connections in residual networks expose the inner block features in such a manner that the fine-tuning process can utilize these features in a near-linear way \citep{evci2022head2toe}. }

Fine-tuning is carried out with a standard stochastic learning procedure (e.g. \textsc{sgd} or \textsc{adam}) after applying a very large dropout to the penultimate layer representation $\Phi$. \LB{Unlike \citep{srivastava2014dropout}, we only apply dropout on the penultimate layer representation $\Phi$, because skip connections in residual networks expose many inner-layer features to the last linear layer, as illustrated by the decomposition of residual networks proposed by \citet{veit2016residual},}
\begin{align}
\label{eq:residual_feature}
    \Phi(x) &= \underbrace{\vphantom{f_1(x)}x}_{\mathclap{\phi_0(x)}} + \underbrace{f_1(x)}_{\phi_1(x)} + \underbrace{f_2(x+f_1(x))}_{\phi_2(x)} + \dots \notag \\ 
    &= \sum_{i\in [0,\dots,l]} \phi_i(x)\,,
\end{align}
where $f_i$ represents the function implemented by the $i$-th residual block, and
\begin{align}
\label{eq:dropout}
  \Phi_{\mathrm{dropout}}(x) = \frac{m(\lambda)}{1-\lambda} \odot \Phi(x) \, ,
\end{align}
where $\odot$ represents the component-wise product and $m(\lambda)$ is a vector of random Bernoulli variables equal to $0$ with probability $\lambda$ and $1$ with probability $1-\lambda$. The additive decomposition of $\Phi(x)$ in equation \eqref{eq:residual_feature} makes clear that applying dropout to $\Phi(x)$ simultaneously blocks the contributions~$\phi_i(x)$ of all residual blocks.

In this work, this approach is called \textbf{very-large dropout}, because the dropout rate ($\sim$90\%) is far larger than people used before.

\section{Experiments}
\label{sec:experiements}

\paragraph{Dataset} 

We perform most experiments using \textsc{pacs} \citep{PACS}, \textsc{vlcs} \citep{vlcs}, \textsc{office\,home} \citep{officehome}, and \textsc{terra\, incognita} \citep{terraincognita} datasets. 
These datasets spam in diverse domains, from wild images with different environment conditions to artificial sketching and painting, from natural animals to home furniture. With $9,991$ to $24,788$ examples, these datasets are substantially smaller than the pre-training dataset \textsc{ImageNet} with 1.2\textsc{m} examples. 
%

Each of these datasets is divided into four sub-datasets that share the same target label categories but follow a different distribution. For example, one sub-dataset of \textsc{pacs} contains simple sketch images of `dog' and `elephant', while another sub-dataset contains real photos of `dog' and `elephant'. This makes it possible to conveniently evaluate \ood. performance by fine-tuning on three sub-datasets and testing on the fourth one. 

\paragraph{Models}

We carry out experiments using two wisely used residual architectures. For the \textbf{convolutional network} experiments, we use a \textsc{ResNet50} architecture \citep{he2016deep} with 25\textsc{m} parameters.\footnote{\url{https://pytorch.org/blog/how-to-train-state-of-the-art-models-using-torchvision-latest-primitives/}} For the \textbf{visual transformer} experiments, we use the large vision transformer \textsc{ViT-L-16} \citep{dosovitskiy2020image} with 304\textsc{m} parameters.\footnote{\url{https://github.com/pytorch/vision/tree/main/references/classification\#vit_l_16}}

\paragraph{Pre-training} 

Unless otherwise stated, all experiments are carried out using networks
pre-trained using refined data augmentations initially introduced in the context of residual networks: \textsc{trivialaugment} \citep{muller2021trivialaugment}, \textsc{cutmix} \citep{yun2019cutmix}, and \textsc{random\,erasings} \citep{zhong2020random}. We use these augmentations to mimic the properties of large foundational models trained using very large and diverse pre-training data.

\paragraph{Baselines} 

Using these same datasets, \citet{gulrajani2020search} argue that plain Empirical Risk Minimization (\textbf{\textsc{erm}}) equals and often betters the \ood. performance of purposefully designed methods, such as \textbf{\textsc{coral}} \citep{sun2016deep}, \textbf{\textsc{dro}} \cite{Sagawa2019DistributionallyRN}, \textbf{\textsc{mldg}} \cite{li2018learning}, \textbf{\textsc{dann}} \cite{Ganin2015DomainAdversarialTO}, \textbf{\textsc{c-dann}} \cite{Li2018DomainGV}, \textbf{\textsc{mmd}} \cite{lidomainadversarial}, \textbf{\textsc{vrex}} \citep{krueger2021out}, and \textbf{\textsc{irm}} \citep{arjovsky2019invariant}.  More recently, \citet{wa}, \citet{swad}, \citet{rame2022diverse}, and \citet{rame2022recycling} find that \textbf{ensemble} and \textbf{weight averaging} methods consistently outperform the \ood. performance of \textbf{\textsc{erm}}. 

Therefore, it is sufficient to compare our results with those of the \textbf{ensemble}, \textbf{weight averaging}, and \textbf{\textsc{erm}} methods which are the strongest available baselines.\footnote{\citet{gulrajani2020search, wa, swad, rame2022diverse, rame2022recycling} provide the details about how ensemble and weighting averaging outperform other baseline methods. }

\begin{table*}[th]
    \caption{\ood. performance comparison between very large dropout, ensembles, and weight averaging methods after hyperparameter selection. The hyperparameter is selected according to the best \iid. performance.}
    \label{tab:diverse_solutions_dominate_2d}
    \medskip
    \centering
    \begin{tabular}{p{3mm}| c|ccc|c|cc}
    \toprule
       & dataset & \textsc{erm} & \makecell{weight average\\ (single run)} & \makecell{ensemble\\ (single run)} & \makecell{very-large \\ dropout} & \makecell{weight average\\(multi run)} & \makecell{ensemble\\(multi run)} \\
    \midrule
       &  \textsc{vlcs} & 78.3 & 79.4 & 79.6 & \textbf{80.1} & 78.8 & 79.1 \\
      &  \textsc{office home} & 71.4 & 72.2 & 72.3 & \textbf{73.6} & 71.3 & 71.3 \\
      \multirow{5}{0pt}{\myrotcell{~~~\textsc{ResNet}}} &  \textsc{pacs} & 87.3 & 86.9 & 87.3 & \textbf{88.5} & 87.0 & 87.1 \\
      &  \textsc{terra incognita} & 51.0 & 53.1 & 52.3 & \textbf{53.9} & 52.0 & 52.5 \\
        \cmidrule{2-8}
     & \textbf{Average} & 72.0 & 72.9 & 72.9& \textbf{74.0} & 72.3 &72.5 \\
     \midrule
     \midrule
    &   \textsc{vlcs} & 78.1 & 78.1 & 77.9 & \textbf{79.0} & 78.4 & 78.4 \\
    &    \textsc{office home} & 74.6 & \textbf{74.8} & \textbf{74.8} & 74.6 & 74.5 & 74.6\\
    \multirow{5}{0pt}{\myrotcell{~~\textsc{ViT-L-16}}} &    \textsc{pacs} & 85.0 & 84.2 & 84.3 & \textbf{86.0} & 84.7 & 84.8 \\
     &   \textsc{terra incognita} & 44.4 & 45.1 & 44.8 & \textbf{45.8} & 44.1 & 44.0\\
        \cmidrule{2-8}
     &   \textbf{Average} & 70.5 & 70.6 & 70.5 & \textbf{71.4} & 70.4 & 70.5\\
    \bottomrule
    \end{tabular}
\end{table*}

\begin{table*}[h]
    \centering
    \caption{Very-large dropout $+$ a $10\times$ larger learning rate in the last layer. The first two columns show that this $10\times$ last-layer learning rate is helpful to \textsc{erm}. Then the middle two columns show that using a large dropout rate vastly improves the \ood. performance of merely using the increased learning rate ($\sim$$1.3\%$). The last two columns reveals that using this $10\times$ larger last-layer training rate yields small or zero incremental improvements over only using a large dropout rate ($\sim$$0.2\%$).}
    \medskip
    \begin{tabular}{c|ccccc}
        \toprule
        dataset & \textsc{erm} & \makecell{10$\times$ last-layer lr} &  \makecell{very-large dropout} & \makecell{very-large dropout\\+ 10$\times$ last-layer lr}\\
        \midrule
        \textsc{vlcs} & 78.3 & 79.9 {\scriptsize(+1.6)} & 80.1 {\scriptsize(+1.8)} & \textbf{80.5 {\scriptsize(+2.2)}} \\
        \textsc{office home} & 71.4 & 71.8 {\scriptsize(+0.4)}  & \textbf{73.6 {\scriptsize(+2.2)}}  & 73.3 {\scriptsize(+1.9)} \\
        \textsc{pacs} & 87.3 & 87.0 {\scriptsize(-0.3)} & \textbf{88.5 {\scriptsize(+1.2)}} &   88.3 {\scriptsize(+1.0)} \\
        \textsc{terra incognita} & 51.0 & 52.2 {\scriptsize(+1.2)} & 53.9 {\scriptsize(+2.9)} & \textbf{54.9 {\scriptsize(+3.9)}} \\
        \midrule
        Average & 72.00 & 72.73 & 74.03 & 74.25 \\
        \bottomrule
    \end{tabular}
    \label{tab:diverse_solutions_dominate_10}
\end{table*}

\subsection{Very large dropout yields better \ood. performance}
\label{sec:dropout_exp}


Table \ref{tab:diverse_solutions_dominate_2d} shows our \textbf{main results} that comparing our very-large dropout approach and baseline methods on four \ood. datasets and two pretrained backbones.
%
\footnote{Code: \url{https://github.com/TjuJianyu/verylarge_dropout}}

\paragraph{{\textsc{erm}}}
results are obtained by fine-tuning \textsc{ResNet50} or \textsc{ViT-L-16} using \textsc{sgd} with 0.9 momentum for $10,000$ iterations.\footnote{We use a batch size $32$ for all \textsc{ResNet} fine-tunings, and reduce the batch size to $16$ for all \textsc{ViT-L-16} fine-tunings due to the \textsc{vram} constraint.} A 10\% learning rate decay is applied at $5000^{th}$ iterations. For each choice of three training sub-datasets, we repeat three experiments for each combination of learning rate in $\{10^{-3}, 5.10^{-4}\}$ and L2 weight decay in $\{10^{-4}, 5.10^{-5}, 10^{-5}\}$. Following \citet{gulrajani2020search}, we prevent overfitting by early-stopping on 20\% hold-out \iid. validation examples, select hyperparameter (for each choice of training sub-datasets) according to the best \iid. performance. Finally, we evaluate the selected models on the fourth sub-dataset and average the four choices of training sub-datasets.

\paragraph{Ensemble (single run)} results are obtained by an ensemble of checkpoints collected (every 300 iterations) along each fine-tuning trajectory. 

\paragraph{Weight average (single run)} results approximate the corresponding ensemble (single run) results by averaging the model weights instead of averaging the model outputs. 

\paragraph{Ensemble (multi run)} results are obtained by an ensemble of final checkpoints collected along all fine-tuning trajectories with different hyper-parameters ($2\times3=6$ in total). 

\paragraph{Weight average (multi run)} results approximate the corresponding ensemble (multi run) results by averaging the model weights. 

\paragraph{Very-large dropout} results are obtained using the same protocol but using a 90\% dropout rate on the penultimate layer representation.

As expected, both ensemble methods \citep{ensemble,dietterich2000ensemble}
and their weight averaging approximation \citep{rame2022diverse,wortsman2022model}
improve \LB{on the} \ood. \textsc{erm} performance. However, fine-tuning with a very large dropout outperforms the \ood. performance of both ensemble and weight averaging methods.

Because \textsc{ResNet50} produces a better performance than \textsc{ViT-L-16} on these \ood. fine-tuning tasks, our experiments in the following sections will be conducted on \textsc{ResNet50}.

\subsection{Very-large dropout $+$ other fine-tuning techniques}
\label{sec:exp_popular_finetuning_techniques}
Various fine-tuning techniques have been proposed to improve the fine-tuning ability to leverage the representations learned by a pre-trained model, such as using a larger learning rate on the last layer \citep{caron2020unsupervised, bardes2021vicreg, kumar2022fine} or, as discussed above, using weight averaging and ensemble methods \citep{rame2022diverse,rame2022recycling,wa}. In this section, we show that incorporating these techniques \emph{in additional to very-large dropout} can further enhance \ood. performance, i.e. very-large dropout approach is compatible to these existing fine-tuning techniques. 

More importantly, very-large dropout approach dominates the \ood. performance improvements. i.e., all these finetuning techniques do not yield much \ood. performance improvements over using large dropout rates alone. 

\begin{table*}[th!]
    \caption{Very-large dropout $+$ ensembles or weight averagings. The \textsc{erm} and very-large dropout results are the same as those reported in Table~\ref{tab:diverse_solutions_dominate_2d}. In contrast, the ensemble and weight averaging results are now obtained by averaging the output or the weights of models fine-tuned \emph{with large dropouts}. Ensemble and weight averaging techniques provide a marginal \ood. performance improvement on \textsc{vlcs} or \textsc{office\,home} and a negligible \ood. performance improvement on \textsc{pacs} or \textsc{terra\,incognita}.}
    \label{tab:incremental_benefits_of_ensemble}
    \medskip
    \centering
    \resizebox{\textwidth}{!}{
    \begin{tabular}{c|cccccc}
    \toprule
       dataset & \textsc{erm} & \makecell{very-large\\dropout} & \makecell{very-large dropout\\+ weight average\\ (single run)} &  \makecell{very-large dropout\\+ ensemble\\ (single run)} & \makecell{very-large dropout\\+ weight average\\ (multi run)} &  \makecell{very-large dropout\\+ ensemble\\ (multi run)} \\
    \midrule
    \textsc{vlcs}& 78.3 & 80.1 & 80.6 & 80.5 & 80.4 & 80.3 \\
    \textsc{office home}&71.4 & 73.6 & 74.2 & 74.3 & 74.4 & 74.2 \\
    \textsc{pacs}& 87.3 & 88.5 & 88.6 & 88.8 & 89.0 & 89.0 \\
    \textsc{terra incognita}& 51.0 & 53.9 & 54.0 & 54.7 & 52.3 & 54.7 \\
    \midrule
    \textbf{Average} &72.0 & 74.0 & 74.4 & 74.6 & 74.0 & 74.6 \\
    \bottomrule
    \end{tabular}
    }
\end{table*}

\subsubsection{Very-large dropout \\$+$ large learning rates for the last layer}

Several authors routinely use a larger training rate on the last layer on the intuition that fine-tuning a pre-trained deep network on a different target task entails training a new last layer from scratch \citep{caron2020unsupervised, bardes2021vicreg, kumar2022fine}.

Table~\ref{tab:diverse_solutions_dominate_10} follows a similar fine-tuning process as in Table~\ref{tab:diverse_solutions_dominate_2d} but uses a $10\times$ larger training rate for the last layer classifier. Comparing the last two columns in Table~\ref{tab:diverse_solutions_dominate_10} shows that incorporating this $10\times$ larger last layer training rate is able to keep or improve the \ood. performance ($\sim$0.2\%). Comparing the middle two columns further shows that using a large dropout rate vastly improves the \ood. performance of merely using the increased learning rate ($\sim$$1.3\%$).

\subsubsection{Very-large dropout \\$+$ ensemble or weight averaging}

Table~\ref{tab:incremental_benefits_of_ensemble} similarly explores the incremental benefits achieved by constructing ensembles or by averaging the weights of models fine-tuned with very large dropouts. The results show that very-large dropout approach is compatible with ensembles and weight averaging apporach to gain a non-negative incremental imporvements in \ood. performance. On the other hand, 
%
comparing Table~\ref{tab:diverse_solutions_dominate_2d} and~\ref{tab:incremental_benefits_of_ensemble} shows that fine-tuning with large dropout rates before computing ensembles or averaging model weights brings large \ood. performance improvements over fine-tuning without dropout. 

In short, \emph{the very-large dropout approach is compatible with other fine-tuning techniques but acts as the leading factor in terms of \ood. performance}.

\subsection{Robustness to hyperparameter selection}
\begin{figure}[th!]
    \centering
\includegraphics[width=0.23\textwidth]{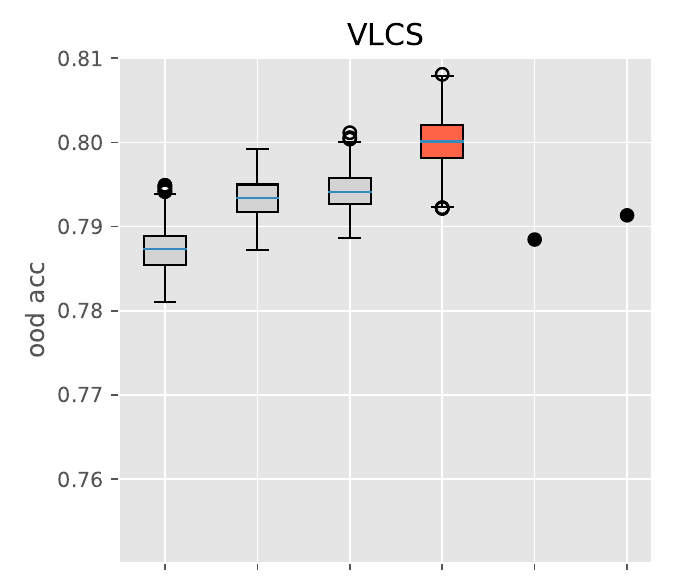}
\includegraphics[width=0.23\textwidth]{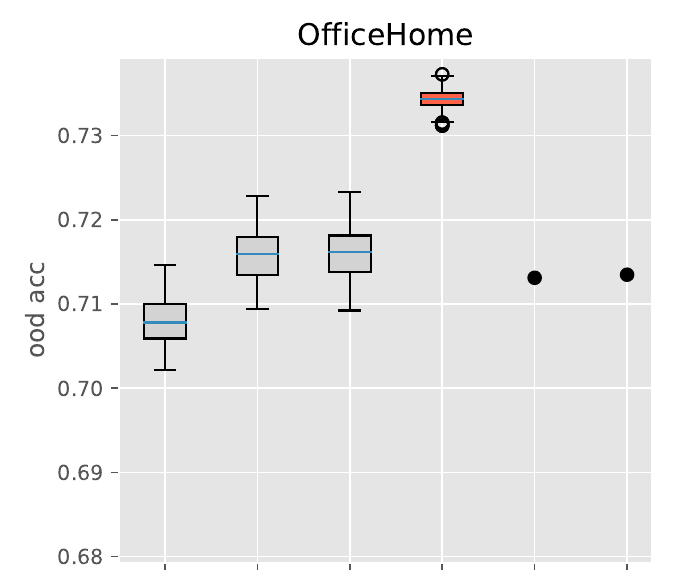} \\ 
\medskip
\includegraphics[width=0.23\textwidth]{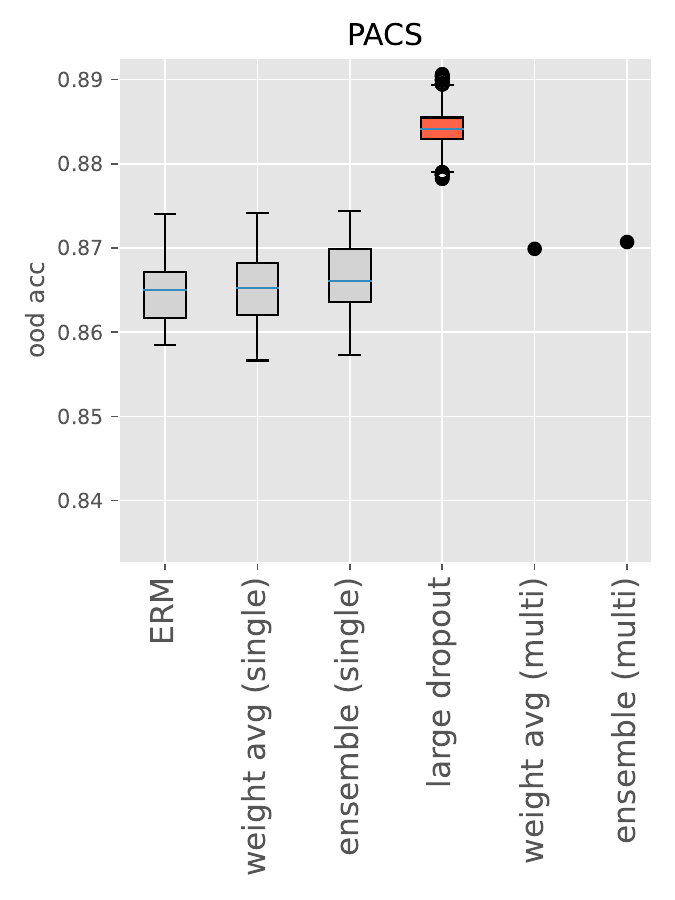}
\includegraphics[width=0.23\textwidth]{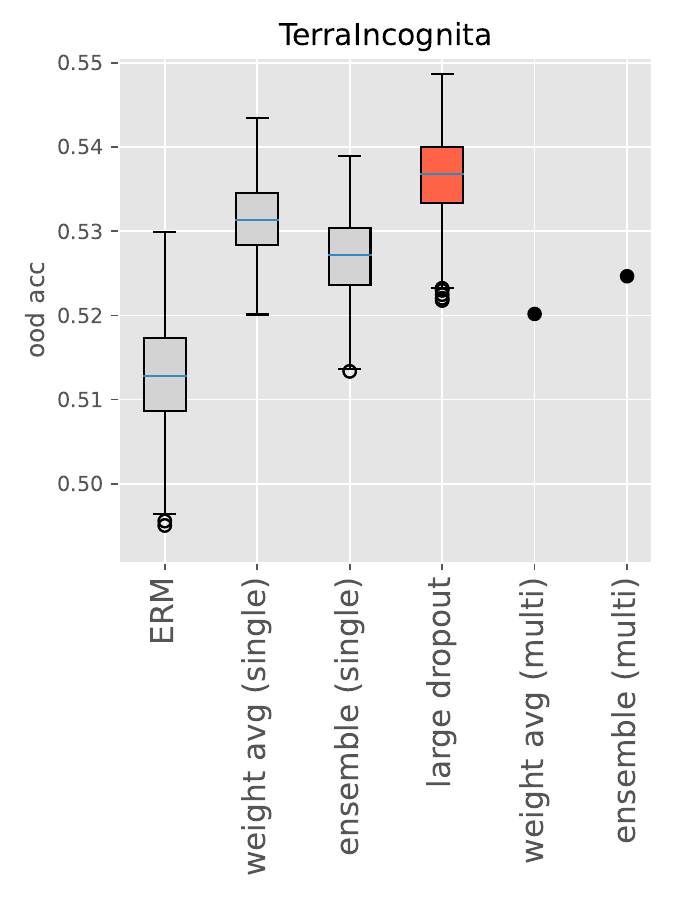} \\
    \caption{\ood. performance comparison between very large dropout, ensembles, and weight averaging methods on four \textsc{DomainBed} tasks. \textbf{\textsc{erm}} results were obtained using plain fine-tuning with different hyperparameters. \textbf{Weight averaging} results either average the model weights collected every 300 iterations along each fine-tuning trajectory or the final model weights of all fine-tuning trajectories as in \citep{rame2022diverse}. \textbf{Ensemble} results average instead the model outputs. Finally, \textbf{large dropout} results were obtained like the \textsc{erm} results but using a 90\% dropout rate on the penultimate layer. Each box summarizes the results obtained with different hyper-parameters combinations.}
    \label{fig:diverse_solutions_dominate_2d}
\end{figure}

Out-of-distribution finetuning performance is known to be sensitive to hyperparameter selection \cite{ahuja2020empirical,wortsman2022model}. To reduce the uncertain of hyperparameter selection, Figure~\ref{fig:diverse_solutions_dominate_2d} presents the box plot of different hyperparameter combinations (where each choice of training sub-datasets searches 6 hyperparameter combinations). 

On all four datasets, the bottom of very-large dropout box (25\% quartile) outperforms the top of other baseline boxes (75\% quartile). On \textsc{office\,home} and \textsc{pacs} datasets, there is even a \emph{large gap} between the worst dropout results and the best baseline results.

%

\subsection{Robustness of dropout rate selection}

To the best of our knowledge, such large dropout rates (90\% and above) are considered unsuitable for training a network from scratch and have not been previously used for fine-tuning either. 
This section study the relationship between dropout rates and \ood. performance. A smooth relationship indicates the robustness of dropout rate selection, while a curly relationship reflects the sensitivity.

Table~\ref{tab:compare_dropout_rate} compares various dropout rates on the four tasks. A $90\%$ dropout rate reliably produces good \ood. performance on all four tasks. The optimal dropout rate for \ood. performance ranges from 90\% to 95\% for \textsc{vlcs} and \textsc{pacs} task (with 10k examples). And becomes slightly smaller, about 90\%, for the slighlty larger datasets \mbox{\textsc{office\,home}} and \mbox{\textsc{terra\,incognita}} (with 15k to 25k examples).

Furthermore, the relationship between dropout rate and \ood. performance are smooth on all four datasets, which makes it easy to select the right dropout rate. 

\begin{table}[]
    \centering
    \caption{Effect of diverse dropout rates during fine-tuning. The best \ood. performances are attained using rates around or above 90\%. A large dropout rate (e.g. 90\%) reliably produces good \ood. performance on all four tasks.}
    \label{tab:compare_dropout_rate}
    \medskip
    \begin{tabular}{c|cccc}
        \toprule
        dropout rate  & 0\% & 50\% & 90\% & 95\% \\
        \midrule
        \textsc{vlcs} & 78.3 & 79.7 & 80.1 & \textbf{80.4} \\
        \textsc{office home} &71.4 & 73.1 & \textbf{73.6} & 73.0 \\
        \textsc{pacs} &87.3 & 88.0 & \textbf{88.5} & 88.4 \\
        \textsc{terra incognita} &51.0 & 52.4 & \textbf{53.9} & 52.3 \\
        \bottomrule
    \end{tabular}
\end{table}

\subsection{When should one apply very-large dropout?}

We have demonstrated that the very-large dropout method delivers consistently better \ood. performance than computing ensembles or weight-averages of models fine-tuned without dropout. However we also have argued that fine-tuning does not create new representations but merely exploits the representations already present in the pre-trained model. Therefore the final \ood. performance of this fine-tuning process must strongly depend on the quality and the diversity of the features present in the pre-trained network (\textit{richer representation}), even if these features are not exploited by the pre-trained network but buried in its hidden layers. i.e. the scope of applying very-large dropout method lies in situations where a \textit{rich representation} has already been established. 

Of course, modern foundational models, where many features are learned from a large and carefully constructed dataset, make this condition relatively easy to achieve. Thus provide a large space to apply this very-large dropout approach. 

In this section, we study this condition precisely. We first study the performance of very-large dropout approach on the scratch-training scenario, where the representation is random. Then we progressively enrich the representation by pretraining and pretraining with enormous augmentations.

\begin{figure}[th!]
\centering
\includegraphics[width=0.33\textwidth]{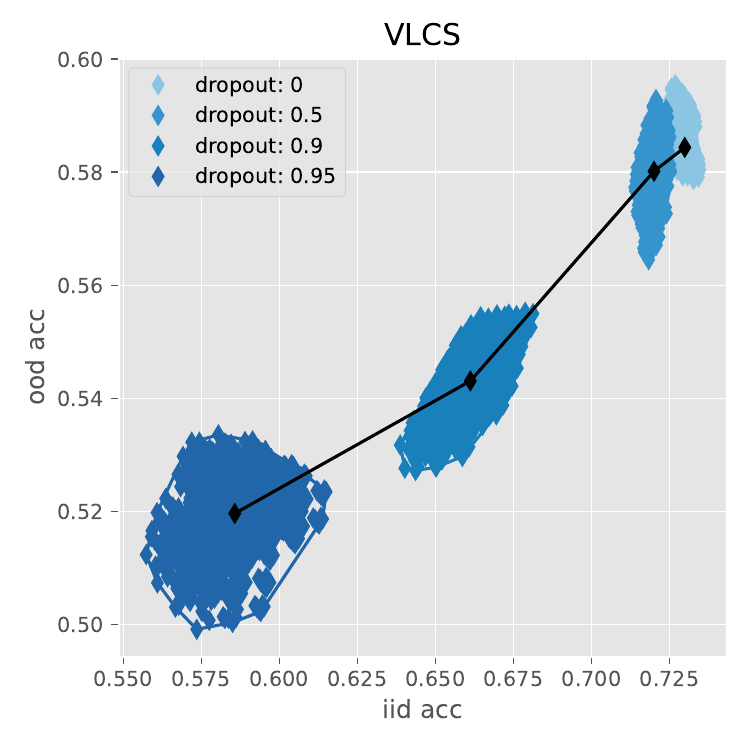}

\caption{Comparison of dropout rates when training a \textsc{ResNet50} network \emph{from scratch} on the \textsc{vlcs} dataset. The optimal dropout rate falls to about zero. Dropout rates greater than 50\% negatively impact both the \iid. and the \ood. performances.\newline }
\label{fig:vlcs_scratch_training}
\end{figure}

\paragraph{{Random initialization and representation.}} Figure~\ref{fig:vlcs_scratch_training} shows the effect of various dropout rates when one trains a network on the \textsc{vlcs} task from scratch, that is starting from a randomly initialized network without pretraining (i.e. random initialization and random representation). The optimal dropout rate falls to about zero. Dropout rates higher than 50\% have a negative impact on both the \iid. and the \ood. performance of the network. \textit{This suggests that high dropout rates make it difficult to create new features (a nonlinear operation), but does not prevent leveraging existing features that were possibly buried in the network inner layers (a linear operation).} This is the idea of richer representation we discussed in section \ref{sec:introduction}.

\paragraph{Richer and richer representation.}
\label{sec:richer_pretrain}
\begin{table*}[th!]
    \centering
        \caption{Comparison of the \ood. performances obtained after fine-tuning two pre-trained networks: \textsc{ResNet\,\#1} and \textsc{ResNet\,\#2}. Hyperparameters are selected according to the best \iid. performance. Compared with \textsc{ResNet\,\#1} \citep{he2016deep}, \textsc{ResNet\,\#2} was pre-trained with the vast array of data augmentation techniques. For each of these two pre-trained networks, we follow two fine-tuning approaches: 1) naive fine-tuning; 2) advanced fine-tuning including various tricks intended to improve the \ood. performance, \eg. large dropout (90\%), weight averaging, and increased last-layer learning rate, using hyper-parameters are selected according to the \iid. performance. Despite all this technology, advanced fine-tuning of a pretrained \textsc{ResNet\,\#1} (2nd column) barely matches the performance of naive fine-tuning on \textsc{ResNet\,\#2} (3rd column). }
    \label{tab:resnetv1v2}
    \medskip
    \begin{tabular}{c|cc|cc}
    \toprule
dataset & \makecell{\textsc{ResNet\,\#1}\\\textsc{erm}} & \makecell{\textsc{ResNet\,\#1}\\{very-large dropout}  } & \makecell{\textsc{ResNet\,\#2}\\\textsc{erm}} & \makecell{\textsc{ResNet\,\#2}\\{very-large dropout}} \\
\midrule
\textsc{vlcs} & 76.7 & 78.1 & 78.3 & 80.1 \\
\textsc{office home} & 68.9 & 69.1 & 71.4 & 73.6 \\
\textsc{pacs} & 86.2 & 86.5 & 87.3 & 88.5 \\
\textsc{terra incognita} & 48.2 & 48.8 & 51.0 & 53.9 \\
\midrule  
\textbf{Average} & 70.0 & 70.6 & 72.0 & 74.0 \\
\bottomrule
    \end{tabular}
\end{table*}

To study the impact of rich representation, we compare the \ood. performance obtained by various methods applied to \textsc{ResNet50} networks pre-trained using the same \textsc{ImageNet} data but using different data augmentation schemes. As explained in the first paragraphs of section~\ref{sec:experiements}, the results reported so far use a network pre-trained using a broad array of data augmentation techniques, termed \textsc{ResNet\,\#2}. We now compare its fine-tuning properties with network termed \textsc{ResNet\,\#1} pre-trained using the simpler protocol described in \citet{he2016deep}.

Table~\ref{tab:resnetv1v2} compares the \ood. performances of both networks after regular fine-tuning and after fine-tuning with very-large dropout. Note that \textsc{ResNet\,\#2} contains richer representations than \textsc{ResNet\,\#1} due to the vast data augmentations. On \textsc{ResNet\,\#1}, where the representation is richer than random representation, a very-large dropout rate (0.9) starts to help \ood. performance (0.6\%). On \textsc{ResNet\,\#2}, where the representation is richer than \textsc{ResNet\,\#1}, the same very-large dropout approach vastly boosts \ood. performance (2\%).

The results in this section showcase an increasing \ood. benefits of the very-large dropout approach as the representation getting richer. Starting from the scale of \textsc{ResNet}50 and \textsc{ImageNet}, the \ood. benefits of a very large dropout becomes significant. 

In the context of large foundational models, both model size and dataset size are far larger than \textsc{ResNet}50 neural network and \textsc{ImageNet} dataset. Thus the space to apply this very-large dropout approach is large.










\section{Discussion}


The \ood. performance of fine-tuning with very large dropout consistently exceeds that achieved by popular techniques such as ensemble and by more recent techniques such as weight averaging.
Furthermore, ensemble and weight averaging techniques only bring a small incremental improvement when applied on top of fine-tuning with large dropout rates. This suggests that very large dropout implements a key factor that favors \ood. performance, which we believe is related to
seeking features of type (a) among features of type (b) as explained in the introduction.

Both ensemble and weight-averaging techniques can be used for training a network from scratch or for fine-tuning a pre-trained network. In contrast, very large dropout rates cannot be realistically used when training a network from scratch. We argue that they work for fine-tuning because fine-tuning is well approximated as a linear process that can leverage their existing or buried features of a pre-trained network but cannot create new ones. Using large dropout rates is akin to a form of L2 regularization, expressing a richer set  of features even if redundant.  

This result also illustrates how the \iid. and \ood. scenarios can call for very different techniques.
It is well known that sparse representations can be very helpful in the \iid. scenario,
and it is increasingly clear that rich representations are preferable in 
the \ood. scenario \citep{zhang2022rich,zhang2023learning,chen2023towards}.
There are no reasons to expect that the many techniques designed for the \iid. scenarios will systematically help \ood. generalization. The very-large dropout case is one of many such examples.

\section*{Impact Statements}
This paper presents work whose goal is to advance the field of Machine Learning. There are many potential societal consequences of our work, none which we feel must be specifically highlighted here.

\clearpage
\bibliography{verylargedropout}
\bibliographystyle{icml2025}

\newpage
\appendix
\onecolumn


\hrule
\begin{center}
\LARGE Fine-tuning with Very Large Dropout
\end{center}

\begin{center}
\large Supplementary Material
\end{center}

\hrule
\vskip 1cm

\section{Experiment details}

\subsection{Training from scratch in Figure~\ref{fig:vlcs_scratch_training}}
The \textsc{vlcs} scratch training experiment in Figure~\ref{fig:vlcs_scratch_training} follows the same pipeline as \ood. fine-tuning experiments. But it uses larger learning rates $\{5.10^{-3}, 10^{-2}\}$ on a random initialized \textsc{ResNet50} network (all weights are trainable).

\subsection{Compute Resources}
\label{apx:computeresource}
All experiments are done on V100 GPUs with Intel(R) Xeon(R) Gold 6230
CPUs. One V100 GPU and less than 32GB RAM are enough to fine-tune one Domainbed dataset within a few hours.

\end{document}